\documentclass[10pt,twocolumn,letterpaper]{article}
\ifdefined\pdfminorversion\pdfminorversion=7\fi
\usepackage[pagenumbers]{cvpr}
\usepackage{multirow}
\usepackage{makecell}
\usepackage{mathtools}
\usepackage{bm}
\usepackage{pifont}
\newcommand{\tit}[1]{\smallbreak\noindent\textbf{#1.}}
\newcommand{\modelname}{\textbf{FAR-Drive}}

\newcommand{\gcmark}{\textcolor{ForestGreen}{\ding{51}}}
\newcommand{\rxmark}{\textcolor{red}{\ding{55}}}

\definecolor{cvprblue}{rgb}{0.21,0.49,0.74}
\usepackage[pagebackref,breaklinks,colorlinks,allcolors=cvprblue]{hyperref}
\hypersetup{pdftitle={FAR-Drive: Frame-AutoRegressive Video Generation for Autonomous Driving},pdfauthor={Yaoru Li, Federico Landi, Marco Godi, Ruiju Fu, Yufei Ma, Muyang Sun, Heyu Si, Mingli Song, Shunyu Liu}}
\title{FAR-Drive: Frame-AutoRegressive Video Generation for Autonomous Driving}
\author{%
Yaoru Li\textsuperscript{1,2}\quad Federico Landi\textsuperscript{2}\quad Marco Godi\textsuperscript{2}\quad Ruiju Fu\textsuperscript{2}\quad Yufei Ma\textsuperscript{2}\\
Muyang Sun\textsuperscript{2}\quad Heyu Si\textsuperscript{1}\quad Mingli Song\textsuperscript{1}\quad Shunyu Liu\textsuperscript{3,*}\\[3pt]
\textsuperscript{1}Zhejiang University\quad \textsuperscript{2}Huawei Technologies Ltd.\\
\textsuperscript{3}Nanyang Technological University\\
{\tt\small liyaoru@zju.edu.cn\quad shunyu.liu@ntu.edu.sg}\\
{\small\textsuperscript{*}Corresponding author.}}
\begin{document}
\maketitle

\begin{abstract}
Driving simulation requires visual observations to reflect updates in camera pose, road layout, and surrounding traffic at each simulation step. Generating multiple frames per update limits the temporal granularity at which new controls can be incorporated. Frame-level autoregressive generation supports such per-frame updates, but requires a model to use visual history effectively and remain stable as generated frames are fed back into subsequent predictions. Each update also requires a new inference pass, making per-frame latency an important consideration. In this work, we propose \modelname, a \textbf{F}rame-level \textbf{A}uto\textbf{R}egressive video generation framework for autonomous driving, which integrates a multi-view diffusion transformer with fine-grained structured control to produce geometrically consistent multi-camera videos. We design a two-stage training strategy: (i) adaptive reference-horizon conditioning, which varies the temporal conditioning window for long-horizon consistency, and (ii) blend-forcing training, which gradually shifts from ground-truth to self-conditioned inputs to improve autoregressive stability. We further integrate system-level inference optimizations to reduce per-frame latency. Experiments on nuScenes show that our method achieves leading reported results in our comparison of autoregressive multi-view driving video generators, with second-level latency on a single GPU.
\end{abstract}

\section{Introduction}
\label{sec:intro}

Autonomous driving simulation is critical for training and evaluating driving systems, as real-world data collection is costly and limited in safety-critical scenarios~\cite{dosovitskiy2017carla}. While public datasets~\cite{sun2020scalability, caesar2020nuscenes, xiao2021pandaset} provide large-scale annotated logs, recorded sensor streams cannot respond to dynamic agent actions. Recent advances in generative video models~\cite{wan, hunyuan, sd3, flux1} offer a promising foundation for learning-based simulation. Closed-loop methods~\cite{yang2025drivearena, mei2024dreamforge, gao2024vista} aim to bridge this gap by enabling agent--environment interaction. Driving simulation requires successive visual observations to reflect updated camera poses, road layouts, and traffic states. This motivates visual generation models that incorporate changing scene conditions.

The temporal granularity of these updates depends on how video frames are generated. Chunk-wise generation jointly models several future frames, providing temporal context within each generated segment. However, generating multiple future frames before incorporating the next control update limits this temporal granularity. Frame-level autoregressive generation allows the structured controls to be updated at every generated frame, offering a finer interface between scene evolution and visual synthesis. Motivated by this finer control granularity, we study how to train a multi-view video model for sustained frame-level autoregressive generation of multi-view driving videos.

\begin{figure*}[t!]
\centering
\includegraphics[width=\textwidth]{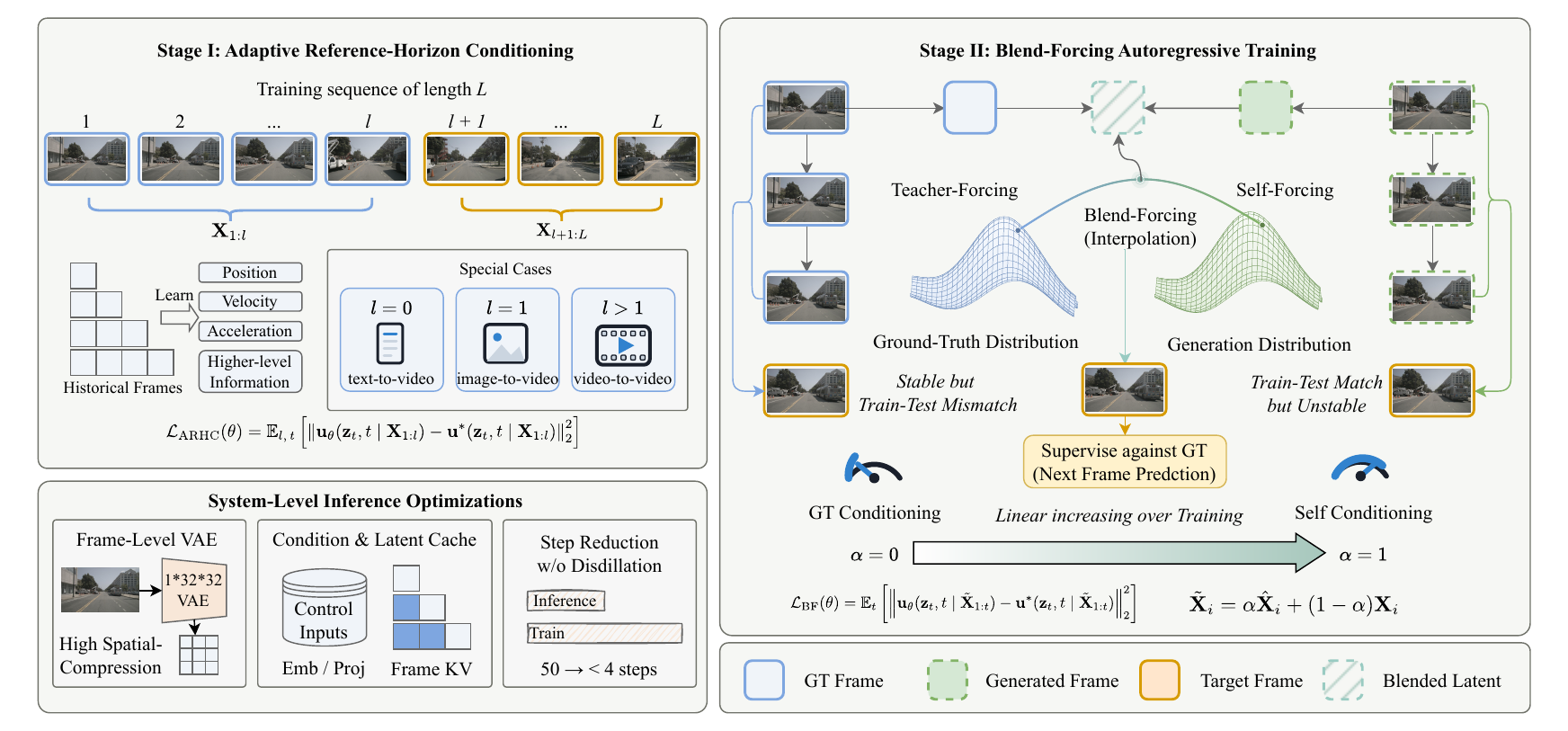}
\caption{
Overview of \modelname~framework for autonomous driving video generation.
To improve long-horizon consistency, we propose adaptive reference-horizon conditioning, which exposes the model to reference inputs of varying lengths during training. From the perspective of video generation, this corresponds to learning different types of generation tasks; in driving scenarios, it amounts to learning to predict environmental evolution from different levels of physical information.
We further propose blend-forcing to enhance autoregressive stability: the interpolated blended latent stays close to the ground-truth distribution in early training to stabilize optimization, and approaches the generation distribution in later training to increase exposure to model-generated history.
In addition, system-level engineering optimizations are performed to achieve low-latency inference during autoregressive rollout.
}
\label{fig:introduction}
\end{figure*}

Long-horizon consistency is a prerequisite for reliable simulation. In frame-level generation, each new frame must remain consistent with the visual history. Multiple historical frames provide motion cues that are ambiguous in a single image, while different reference lengths expose different amounts of temporal information. Training with a fixed reference horizon does not explicitly prepare the model to use these different histories. A second challenge concerns the reliability of the history itself: during autoregressive rollout, generated frames become inputs to subsequent predictions, allowing small errors to accumulate. Self-conditioning strategies such as Self Forcing~\cite{selfforcing} address the train--test distribution mismatch by incorporating model-generated frames during training. Self Forcing first initializes a causal generator through ODE-based finetuning and then performs autoregressive self-rollout during post-training. During these rollouts, prediction errors become part of the conditioning history for subsequent frames. This motivates a gradual transition to generated history: rather than exposing the model to fully generated references from the outset of autoregressive training, we progressively introduce generated content while retaining time-aligned ground-truth guidance for subsequent frame prediction.

In addition to these modeling challenges, practical deployment further imposes low-latency constraints. Each frame-level update requires a new inference pass, making repeated processing of historical frames and structured conditions costly. Although diffusion distillation techniques such as distribution matching distillation~(DMD)~\cite{dmd, dmd2} improve efficiency, frame-level autoregressive generation also exhibits intrinsic redundancies beyond what diffusion distillation alone addresses. Achieving long-horizon consistency, autoregressive stability and low-latency inference speed thus requires coordinated design across training strategy and system-level inference optimization.

Motivated by these observations, we propose \modelname, a frame-level autoregressive video generation framework for autonomous driving, as shown in Fig.~\ref{fig:introduction}. Built upon a controllable multi-view MMDiT~\cite{dit,sd3}, our framework adopts a two-stage training strategy. We introduce adaptive reference-horizon conditioning, which varies the temporal conditioning window during training. To improve autoregressive stability, we devise a blend-forcing strategy that gradually interpolates between ground-truth and generated frames during training, improving robustness while preserving generation stability. Furthermore, we incorporate a high spatial-compression autoencoder~(AE)~\cite{chen2024deep} alongside control state and latent caching mechanisms to significantly accelerate inference. Experiments on nuScenes evaluate generation quality, long-horizon autoregressive stability, and inference efficiency. Our contributions are:
\begin{itemize}

\item We propose \modelname, a frame-level autoregressive video generation framework for autonomous driving built upon a controllable multi-view diffusion transformer backbone, which jointly addresses long-horizon consistency, autoregressive stability, and low-latency inference through coordinated training and inference design.

\item We introduce a novel two-stage training strategy including (i) adaptive reference-horizon conditioning to enhance long-horizon consistency, and (ii) blend-forcing training to improve autoregressive stability while maintaining robustness under self-conditioning.

\item We develop a set of inference optimizations to significantly reduce rollout latency for autoregressive generation as an engineering contribution.

\item Extensive experiments demonstrate superior performance, validating the effectiveness of our framework for autoregressive autonomous driving video generation.

\end{itemize}

\section{Related Work}
\label{sec:rw}

\textbf{Conditional Video Generation} has recently emerged as a promising paradigm for autonomous driving~\cite{gao2023magicdrive,yang2023bevcontrol,swerdlow2024street,drivefix}, enabling the synthesis of future visual observations conditioned on scene context, sensor inputs, or structured representations. One line of research leverages volumetric or occupancy-based intermediate representations for controllable multi-view generation~\cite{yang2025x,li2025occscene,lu2024wovogen,li2025uniscene,guo2025dist}. Other works explore single-view driving generation~\cite{kim2021drivegan,wang2023drivedreamer,chen2025drivinggpt,hu2024drivingworld,gao2024vista,zhang2025epona}, while multi-view methods incorporate structured scene representations and cross-view interaction~\cite{zheng2024genad,wen2024panacea,wang2024driving,zhao2025drivedreamer,gao2025magicdrive,li2025omninwm}. MaskGWM~\cite{Ni_2025_CVPR} combines diffusion generation with video mask reconstruction and row-wise cross-view modeling, with variants for long-horizon and multi-view generation. DriveScape~\cite{Wu_2025_CVPR} uses bidirectional modulated transformers to fuse multi-view features with spatial conditions for high-resolution driving video generation. Our work builds on structured multi-view conditioning and studies how to train the generator for sustained frame-level autoregressive rollout over long sequences.

\smallbreak\noindent\textbf{Autoregressive Video Generation} has recently attracted increasing attention, particularly for world models as simulators in autonomous driving~\cite{hu2023gaia,yang2025drivearena,gao2024vista,mei2024dreamforge,chen2024unimlvg}. These models generate successive observations conditioned on visual history and external inputs. Driving simulators such as DriveArena~\cite{yang2025drivearena} connect scene-state updates with conditional video generation. When multiple future frames are generated before the next condition update, the update granularity is tied to the generated chunk. We focus on frame-level generation, where each generation step incorporates the structured controls associated with the next frame. Training for this setting requires attention to both the reference horizon and exposure to generated inputs. Fixed-horizon conditioning in driving video models~\cite{wen2024panacea,zhao2025drivedreamer} motivates our ARHC strategy, which varies the amount of clean reference history during training. Scheduled sampling~\cite{bengio2015scheduled} progressively replaces ground-truth inputs with model predictions to address exposure bias. Self Forcing~\cite{selfforcing} performs autoregressive self-rollout and applies video-level distribution-matching objectives; Self-Forcing++~\cite{selfforcing++} extends supervision to sampled segments of self-generated long videos. Our BF strategy blends generated and time-aligned ground-truth latents continuously, increasing the generated contribution over training. Together, ARHC and BF address the length and reliability of the visual history used for frame-level driving video generation.

\section{FAR-Drive Framework}
\label{sec:method}

We design a unified framework aiming to enable fine-grained, frame-level autoregressive video generation for autonomous driving, which integrates (i) a controllable multi-view diffusion transformer~(MMDiT) that combines coherent multi-camera generation with structured driving condition injection, (ii) a two-stage training strategy that improves long-horizon consistency while enhancing autoregressive stability, and (iii) system-level inference optimizations that reduce latency for frame-level generation, which is important for fine-grained interactive deployment.

\subsection{Problem Definition}
\label{subsec:problem_def}

We generate synchronized multi-view observations under frame-specific scene controls. Let $\mathbf{I}_i=\{\mathbf{I}_i^{(v)}\}_{v=1}^{V}$ denote the RGB images at frame index $i$, with $\mathbf{I}_i^{(v)}\in\mathbb{R}^{H\times W\times3}$. A frame-wise encoder $E$ maps them to clean latents $\mathbf{X}_i=E(\mathbf{I}_i)$, and $D$ denotes its decoder. We retain $\mathbf{X}$ for latent sequences throughout the framework figures and training objectives.
The reference index set $\mathcal{H}_i\subseteq\{1,\ldots,i\}$ selects the available visual history; its length varies during training and is configurable at inference.

The controls for target frame $i$ are $\mathbf{C}_i=\{\mathbf{P}_i,\mathbf{B}_i,\mathbf{E}_i,\mathbf{M}_i,\mathbf{c}\}$: camera intrinsics and extrinsics $\mathbf{P}_i$, object 3D boxes $\mathbf{B}_i$, an ego-centric BEV map $\mathbf{E}_i$, the ego-pose transformation $\mathbf{M}_i\in\mathbb{R}^{4\times4}$ relative to the first frame, and a scene caption $\mathbf{c}$. The next-frame latent is generated as
\begin{equation}
\hat{\mathbf{X}}_{i+1}=\mathcal{F}_\theta(\mathbf{X}_{\mathcal{H}_i},\mathbf{C}_{i+1};\xi_{i+1}),
\end{equation}
where $\xi_{i+1}$ denotes sampling noise; decoding gives $\hat{\mathbf{I}}_{i+1}=D(\hat{\mathbf{X}}_{i+1})$. Generated latents enter the reference history for subsequent steps. In our experiments, controls are constructed from the nuScenes scene records and supplied as a fixed sequence; generated images do not modify these controls. This interface permits per-frame control updates while preserving temporal causality in visual generation.

\begin{figure}[t!]
\centering
\includegraphics[width=0.89\columnwidth,trim=20bp 21bp 22bp 20bp,clip]{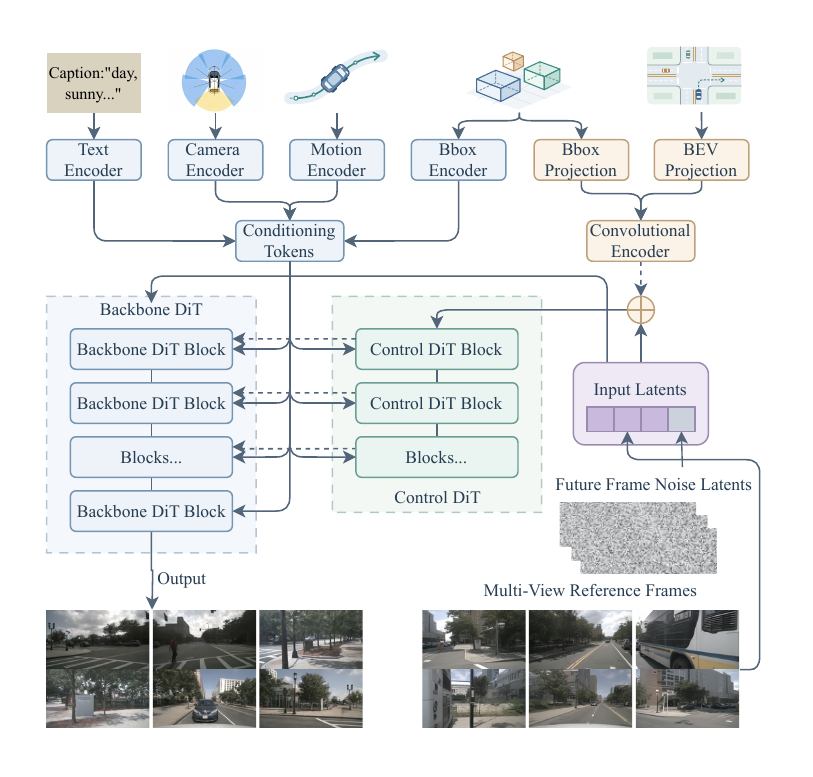}
\caption{Overview of the proposed multi-view MMDiT architecture.
Reference frames and noise latents are fed into the backbone DiT while
structured controls including projected 3D bounding boxes and BEV maps are first encoded by a convolutional encoder and then processed by a control DiT conditioned on the same latent inputs.
All control signals are additionally aggregated into conditioning tokens, which are injected into every layer of both the backbone and control DiTs.
The outputs of each Control DiT block are injected into the corresponding backbone DiT blocks through zero-initialized projection layers (dashed arrows).
Since the Control DiT has fewer layers than the backbone, control injection is applied only to the early backbone blocks.}
\label{fig:method}
\end{figure}

\subsection{Controllable Multi-view DiT Architecture}
\label{subsec:multiview_dit}

To enable temporally consistent, geometrically aligned, and structurally controllable multi-view video generation for autonomous driving, we design a controllable multi-view MMDiT architecture as shown in Fig.~\ref{fig:method}, which explicitly conditions video generation on diverse control signals and reference frames. The framework supports multiple controllable dimensions, allowing each generated frame to reflect its updated structured controls.

\tit{MMDiT Backbone} We adopt a text-image-video-to-video~(TIV2V) MMDiT as the backbone. To support autoregressive generation, the model operates with causal masking along the temporal dimension, ensuring that each frame attends only to its current and preceding frame tokens during training and inference.

\tit{Multi-view Attention} On top of the backbone, we apply sequential token reshaping to repurpose temporal self-attention for cross-view interaction across all camera views at each timestep, similar to DIVE~\cite{jiang2024dive}. Temporal attention operates over the spatial and temporal tokens of each camera stream independently, while cross-view attention operates over the spatial tokens of all views at the same timestep. This arrangement combines causal temporal modeling with synchronized cross-view interaction. We insert one cross-view attention block after every four backbone blocks, forming seven backbone units from 28 MMDiT blocks. Each cross-view block is initialized from the preceding causal video block and uses a zero-initialized output projection. The backbone has a hidden dimension of 1792 and 14 attention heads; the shallower control branch contains three units. The full model has approximately 3.7B parameters, including 2.5B in the backbone, 0.3B in cross-view modules, 0.8B in the control branch, and 0.1B in auxiliary modules. Appendix~\ref{sec:app:model} provides the detailed block arrangement.

\tit{Control Encoders}
We adopt a control processing design similar to MagicDrive~\cite{gao2023magicdrive}.
Text, camera parameters, ego-motion, and 3D bounding boxes~(Bboxes) are first encoded by modality-specific encoders and fused into conditioning tokens.
In parallel, spatial control signals, including projected Bboxes and BEV maps, are rendered into image-aligned representations and processed by a convolutional encoder before being fed into the control DiT.

\tit{DiT-based Control}
To inject structured control signals into video generation in a fine-grained manner, we adopt a dual-MMDiT architecture inspired by ControlNet-style conditioning~\cite{chen2024pixart}. The model consists of a backbone MMDiT $\mathcal{B}$ responsible for primary video generation and a shallower control MMDiT with the same block design $\mathcal{C}$ dedicated to processing external control inputs (\eg, road layouts and bbox sketches). While both branches use the same block design, they maintain separate parameters during training. Let $x^{(l)}$ and $c^{(l)}$ denote the hidden states of the backbone and control branches at layer $l$, respectively, and let $u$ represent the encoded control signal. The control branch is initialized by injecting the control signal through a zero-initialized projection $\mathrm{Proj}_l(\cdot)$ and adding it to the backbone input:
\begin{equation}
\begin{aligned}
c^{(0)} &= x^{(0)} + \mathrm{Proj}_{\mathrm{in}}(u), \\
c^{(l+1)} &= \mathcal{C}^{(l)}(c^{(l)}), \\
x^{(l+1)} &= \mathcal{B}^{(l)}(x^{(l)}) + \mathrm{Proj}_l(c^{(l+1)}).
\end{aligned}
\end{equation}

Injection applies only to the backbone blocks paired with control blocks; later backbone blocks have no control residual. Parameter counts and block depths are given in Appendix~\ref{sec:app:model}. In this design, each control block produces two effects: its standard output is propagated to the next control block, while a zero-initialized residual projection of the same hidden state is injected into the corresponding backbone block. The zero initialization ensures that, at the beginning of training, the control branch does not perturb the pretrained backbone behavior. As training progresses, the control pathway gradually learns to modulate generation in a layer-wise manner, enabling precise structural conditioning while preserving the generative capacity and stability of the backbone MMDiT.

\tit{Dense road layout and bbox sketch projections}
To construct spatially explicit control inputs for multiview generation, we follow a layout projection strategy similar to DriveArena~\cite{yang2025drivearena}. Specifically, we explicitly project the BEV road layout and 3D object bounding boxes into each camera view based on known camera intrinsics and extrinsics. These projections produce per-view layout canvases that provide dense spatial guidance for lane structure and object placement. The projections are concatenated on channels and encoded with a causal 3D convolutional encoder before entering the control MMDiT.

\subsection{Two-Stage Training Strategy}
\label{subsec:training}
We use two training stages: adaptive reference-horizon conditioning (ARHC), followed by Blend-Forcing (BF). To distinguish video time from diffusion time, $i$ denotes a frame index and $\tau\in[0,1]$ denotes flow time. For a clean target latent $\mathbf{X}_i$ and noise $\epsilon_i\sim\mathcal{N}(0,I)$, a general differentiable interpolation path is
\begin{equation}
\begin{aligned}
\mathbf{Z}_{i,\tau}&=a(\tau)\mathbf{X}_i+b(\tau)\epsilon_i,\\
\mathbf{U}^{*}_{i,\tau}&=a'(\tau)\mathbf{X}_i+b'(\tau)\epsilon_i,
\end{aligned}
\end{equation}
where $a(0)=0$, $b(0)=1$, $a(1)=1$, and $b(1)=0$. This notation specifies the path and velocity target without committing to a particular noise schedule. For target indices $\mathcal{T}$, reference history $\mathcal{R}$, and controls $\mathbf{C}$, the conditional flow-matching objective~\cite{flowmatching} is
\begin{equation}
\begin{split}
\mathcal{L}_{\mathrm{FM}}(\mathcal{T},\mathcal{R})=\mathbb{E}\Bigg[\frac{1}{|\mathcal{T}|}\sum_{i\in\mathcal{T}}
\Big\|\mathbf{u}_{\theta,i}(\mathbf{Z}_{\mathcal{T},\tau},\tau\mid\mathcal{R},\mathbf{C})\\
-\mathbf{U}^{*}_{i,\tau}\Big\|_2^2\Bigg].
\end{split}
\end{equation}
The expectation covers data, noise, flow time, and sampled reference horizons. The attention mask determines which target-frame tokens can interact.

\tit{Stage I: Adaptive Reference-Horizon Conditioning}
ARHC samples a reference length $l\in\{0,\ldots,L-1\}$ for a training clip of length $L$ and conditions on the clean prefix to predict the remaining targets:
\begin{equation}
\mathcal{L}_{\mathrm{ARHC}}=\mathbb{E}_{l\sim p(l)}\left[\mathcal{L}_{\mathrm{FM}}(\{l+1,\ldots,L\},\mathbf{X}_{1:l})\right].
\end{equation}
The cases $l=0$, $l=1$, and $l>1$ provide no visual history, one reference frame, and multiple reference frames, respectively; structured scene controls are retained in all cases. Varying the history encourages the model to use temporal cues across different conditioning horizons, in contrast to fixed-horizon conditioning~\cite{wen2024panacea,zhao2025drivedreamer}. The distribution $p(l)$ can be adapted to the intended rollout setting; our short-horizon-biased sampling is described in Appendix~\ref{app:train}.
Both bidirectional pretraining and causal finetuning use ARHC, with the temporal attention mask distinguishing these two phases.

\tit{Stage II: Blend-Forcing Autoregressive Training}
Inspired by self-forcing~\cite{selfforcing} and resample-forcing~\cite{resampleforcing}, BF introduces generated history during training to address exposure bias~\cite{bengio2015scheduled}. Starting from a ground-truth initial frame, the model generates successive latents under the corresponding scene controls. Each generated latent is blended with its time-aligned ground-truth counterpart:
\begin{equation}
\tilde{\mathbf{X}}_i=\alpha\hat{\mathbf{X}}_i+(1-\alpha)\mathbf{X}_i.
\end{equation}
The blended history conditions subsequent predictions. For the supervised frame indices $\mathcal{S}$, the objective is
\begin{equation}
\begin{split}
\mathcal{L}_{\mathrm{BF}}=\mathbb{E}\Bigg[\frac{1}{|\mathcal{S}|}\sum_{i\in\mathcal{S}}
\Big\|\mathbf{u}_{\theta}(\mathbf{Z}_{i,\tau},\tau\mid\tilde{\mathbf{X}}_{\mathcal{H}_{i-1}},\mathbf{C}_i)\\
-\mathbf{U}^{*}_{i,\tau}\Big\|_2^2\Bigg].
\end{split}
\end{equation}
We roll out up to 16 frames and apply this loss to the last eight frames. Each BF training step is paired with a standard flow-matching update to maintain generation quality. The coefficient increases as $\alpha_k=\min(10^{-4}k,1)$, reaching approximately $0.7$ at the reported 7,000-step checkpoint. Thus training progressively increases exposure to generated history while retaining a ground-truth contribution.

\tit{Joint Supervision on Clean and Blended Histories}
Let $h$ concatenate the clean reference latents and $h+e$ their time-aligned generated counterparts. The blended history is $h+\alpha e$, so its deviation from the clean history scales continuously with $\alpha$. For a fixed noisy target, flow time, and scene controls, define
\begin{equation}
f(s)=\mathbf{u}_\theta(\mathbf{Z}_{i,\tau},\tau\mid h+se,\mathbf{C}_i),
\quad r_s=f(s)-\mathbf{U}^{*}_{i,\tau}.
\end{equation}
The residuals $r_0$ and $r_\alpha$ correspond to clean-history FM and blended-history BF supervision, respectively, while $r_1$ describes prediction under generated history. Assume the predictor's Jacobian with respect to history is $M$-Lipschitz along the segment $\{h+se:s\in[0,1]\}$. For $0<\alpha\le1$, this local smoothness yields
\begin{equation}
\begin{split}
\|r_1\| \le {}& \frac{1}{\alpha}\|r_\alpha\|
+\frac{1-\alpha}{\alpha}\|r_0\|\\
&+\frac{M}{2}(1-\alpha)\|e\|^2.
\end{split}
\label{eq:bf_transfer}
\end{equation}
The first two terms connect generated-history prediction to the residuals supervised by the paired training objectives; the final term captures curvature along the history segment. Small $\alpha$ provides a gentle introduction to generated-history perturbations. As $\alpha$ increases, the blended supervision moves closer to generated-history conditioning, and both extrapolation coefficients and the curvature term decrease. Standard FM maintains prediction quality at the clean endpoint, while BF trains the response along model-generated error directions. This analysis motivates the combination of clean-history supervision and a progressively increasing generated contribution. The full derivation is given in Appendix~\ref{app:blend_theory}.

\subsection{System-Level Inference Optimizations}
\label{subsec:infer}

To meet the low-latency requirements of frame-level autoregressive generation, we incorporate system-level optimizations to accelerate inference during autoregressive rollout.

\tit{Frame-Level Autoencoder Design}
We adopt a DC-AE with spatial-only compression ($1\times32\times32$)~\cite{chen2024deep}, which avoids the temporal entanglement of standard $4\times8\times8$ video VAEs and enables efficient frame-level generation.

\tit{Sampling Step Reduction}
The final model supports few-step sampling without additional distillation. Table~\ref{tab:inference_ablation} evaluates one to twenty sampling steps, allowing generation quality and latency to be selected according to the intended application.

\tit{Latent Caching}
During autoregressive rollout, we adopt a KV-cache mechanism for reference frames. Their encoded latents and attention keys/values are reused across subsequent steps, with only the newly generated frame contributing fresh queries, which avoids redundant encoding and repeated attention computation over historical frames.

\tit{Condition Caching}
Unlike standard diffusion pipelines where condition encoding (\eg, text prompts) is lightweight, our structured control inputs introduce non-trivial encoding overhead. Since the control state remains fixed throughout the denoising trajectory, we encode structured control inputs (\eg, road layouts, object projections) once and cache the features across all diffusion steps, significantly reducing inference time without affecting quality.

\section{Experiments}
\label{sec:exp}

We conduct a series of experiments to systematically evaluate the proposed \modelname~framework, aiming to answer the following questions:
(1) Can frame-level autoregressive modeling maintain strong performance compared with existing driving video generation methods? (Sec.~\ref{main_results})
(2) Can the proposed blend-forcing autoregressive training strategy enable stable and consistent long-horizon rollout? (Sec.~\ref{ablation})
(3) Can the model operate under low-latency constraints while preserving generation quality? (Sec.~\ref{ablation})

\subsection{Experimental Setup}
\label{sec:exp_setup}

\tit{Data} We train on the official nuScenes~\cite{caesar2020nuscenes} training split (700 scenes) and evaluate on the official validation split (150 scenes). Each scene provides six camera streams and scene metadata from which the structured inputs are constructed.

\tit{Model} We adopt a pretrained MMDiT backbone~\cite{sd3}, extended for multi-view video as described in Sec.~\ref{subsec:multiview_dit}. Stage I uses ARHC in both bidirectional and causal phases; the causal mask allows a target frame to attend to its own tokens and preceding frames.

\tit{Training Configuration}
We use 16-frame training clips with six camera views, resizing each image to $576\times1024$. The spatial autoencoder produces an $18\times32$ latent feature map per view. Table~\ref{tab:training_schedule} summarizes the three optimization phases within our two-stage strategy. Unless otherwise specified, the batch size is 64 and the learning rate is $8\times10^{-5}$. Stage I samples reference lengths from 0 to 15, with higher probability assigned to short histories of one to three frames. During Stage II, each BF step is paired with an additional standard flow-matching update. The reported 7,000 BF steps refer to the BF updates, each accompanied by this standard update. The blending coefficient reaches approximately $0.7$ at this checkpoint.

\begin{table}[t]
\centering
\small
\setlength{\tabcolsep}{3pt}
\begin{tabular*}{\columnwidth}{@{\extracolsep{\fill}}llll@{}}
\toprule
\textbf{Phase} & \textbf{Attention} & \textbf{Training} & \textbf{Steps} \\
\midrule
Stage I-a & Bidirectional & ARHC & 10,000 \\
Stage I-b & Causal & ARHC & 20,000 \\
Stage II & Causal & BF + FM & 7,000 \\
\bottomrule
\end{tabular*}
\caption{Training schedule. Each Stage II step consists of one BF update and one standard flow-matching (FM) update.}
\label{tab:training_schedule}
\end{table}

\tit{Metrics}
We evaluate Fr\'echet Inception Distance (FID)~\cite{fid},
Fr\'echet Video Distance (FVD)~\cite{fvd}, mean Average Precision (mAP), and
mean Intersection-over-Union (mIoU).
FID measures frame-level distribution alignment in a deep feature space,
while FVD extends this formulation to video features and captures temporal coherence.
Object-mAP evaluates semantic consistency of detected dynamic objects,
and Map-mIoU quantifies structural alignment in BEV space.
We compute mAP and mIoU following the ECCV CODA workshop protocol~\cite{w-coda}.

\tit{Long-Horizon Evaluation}
For a rollout length $n$, FID is computed from the generated frame at step $n$ and the corresponding real frames across evaluation scenes. FVD is computed using generated and real video sequences of length $n$. Thus the FID curve measures endpoint frame quality and the FVD curve measures video quality over each evaluated horizon. Evaluation extends from 16 to 229 frames, with structured controls following the recorded scene sequence throughout each rollout.

\subsection{Main Results}
\label{main_results}

Table~\ref{tab:video_01} compares our method with prior approaches under different autoregressive (AR) generation units. Existing MagicDrive~\cite{gao2023magicdrive,gao2024magicdrive3d,gao2025magicdrive} series models exhibit limited structural fidelity even when increasing the AR unit. MagicDrive-V2 improves temporal coherence with a long generation horizon, but still lags behind in object- and map-level consistency. DreamForge~\cite{mei2024dreamforge} built on a DiT backbone achieves stronger structural metrics, yet remains inferior to our approach.
Our method achieves the best reported values among the methods listed in Table~\ref{tab:video_01} across all three metrics. Notably, even under strict frame-level autoregressive generation (AR unit = 1) and in a few-step setting (sampling steps = 3), it surpasses prior methods that operate on longer generation units.
Under single-shot 16-frame generation, the model achieves the best overall performance with a larger temporal window and 30 diffusion sampling steps.
Overall, the results validate that our model maintains robust performance even under frame-level autoregressive generation, a capability that is critical for reliable long-horizon video generation.

\begin{table}[t]
\centering
\fontsize{8}{10}\selectfont
\setlength{\tabcolsep}{.3em}
\begin{tabular*}{\columnwidth}{@{\extracolsep{\fill}}lcccc@{}}
\toprule
\textbf{Method} & \makecell[c]{\textbf{AR Unit}} & \textbf{FVD} $\downarrow$ & \textbf{mAP} $\uparrow$ & \textbf{mIoU} $\uparrow$ \\
\midrule
MagicDrive~\cite{gao2023magicdrive} & 60 & 217.94 & 11.49 & 18.27 \\
MagicDrive~\cite{gao2023magicdrive} & 16 & 218.12 & 11.86 & 18.34 \\
MagicDrive3D~\cite{gao2024magicdrive3d} & 16 & 210.40 & 12.05 & 18.27 \\
MagicDrive-V2~\cite{gao2025magicdrive} & 241 & 94.84 & 18.17 & 20.40 \\
DreamForge-DiT~\cite{mei2024dreamforge} & 5 & 103.61 & 19.17 & 34.36 \\
DreamForge-SD1.5~\cite{mei2024dreamforge} & 5 & 233.20 & 22.52 & 32.98 \\

\midrule

\modelname~\textit{(AR Gen.)} & \textbf{1} & \underline{94.69} & \underline{27.39} & \underline{36.35} \\
\modelname~\textit{(Single-shot)} & 16 & \textbf{84.71} & \textbf{28.58} & \textbf{38.62} \\
\bottomrule
\end{tabular*}
\caption{Quantitative comparison on multi-view autonomous driving video generation under different autoregressive (AR) units. AR Unit denotes the number of frames generated per model inference: AR Unit = 1 corresponds to strict frame-level autoregressive rollout, while larger values indicate chunk-wise AR generation.
Our methods generate 16 frames so AR Unit = 16 corresponds to single-shot generation.
For AR generation we use 3 sampling steps per frame, while single-shot generation uses 30 sampling steps.
}
\label{tab:video_01}
\end{table}

Table~\ref{tab_video} compares existing methods under different design choices, including volumetric supervision, multi-view support, video generation, and whether they operate in a first-frame condition-free manner.
Our model satisfies all these properties while maintaining competitive performance in both FID and FVD. For this comparison, we use a 20-step autoregressive model as it provides the best performance for image and video quality.
Among the non-volumetric, multi-view, first-frame-conditioned entries in Table~\ref{tab_video}, our single-shot model outperforms DriveDreamer-2 in FID and FVD and is surpassed by OmniNWM~\cite{li2025omninwm}, whose model has 11B parameters compared with our 3.7B.
A real first frame anchors appearance and scene initialization. Removing it imposes a stricter requirement on visual initialization while retaining the same structured controls: generation starts from noise without visual history.
Baseline values in Tables~\ref{tab:video_01} and~\ref{tab_video} are reported results from prior publications, rather than measurements from retrained baselines under our data split.

\begin{table}[!t]
\centering
\scriptsize
\setlength{\tabcolsep}{1pt}
\begin{tabular*}{\columnwidth}{@{\extracolsep{\fill}}lcccccc@{}}
\toprule
\textbf{Method} & \textbf{Vol. Free} & \textbf{MV} & \textbf{Vid.} & \textbf{FF Free} & \textbf{FID} $\downarrow$ & \textbf{FVD} $\downarrow$ \\
\midrule

X-Scene~\cite{yang2025x} & \rxmark &  \gcmark &  \rxmark & \gcmark & 11.29 & -  \\
OccScene~\cite{li2025occscene} & \rxmark & \gcmark &\gcmark & \gcmark & 11.87  & - \\
WoVoGen~\cite{lu2024wovogen} & \rxmark &  \gcmark & \gcmark & \gcmark & 27.60 & 417.70 \\
UniScene~\cite{li2025uniscene} & \rxmark  & \gcmark & \gcmark & \gcmark & 6.45 & 71.94 \\
DiST-4D~\cite{guo2025dist} & \rxmark & \gcmark & \gcmark& \gcmark & 7.40 & 25.55 \\

\midrule

DriveGAN~\cite{kim2021drivegan} & \gcmark & \rxmark & \gcmark & \gcmark & 73.40 & 502.30 \\
DriveDreamer~\cite{wang2023drivedreamer} & \gcmark & \rxmark & \gcmark & \gcmark & 26.80 & 353.20 \\
DriveDreamer~\cite{wang2023drivedreamer} & \gcmark & \rxmark & \gcmark & \rxmark & 14.90 & 340.80 \\
DrivingGPT~\cite{chen2025drivinggpt} & \gcmark & \rxmark & \gcmark & \gcmark & 12.78 & 142.61  \\
DrivingWorld~\cite{hu2024drivingworld} & \gcmark & \rxmark & \gcmark & \gcmark & 7.40 & 90.90 \\
Vista~\cite{gao2024vista} & \gcmark & \rxmark & \gcmark & \gcmark & 6.90 & 89.40 \\
Epona~\cite{zhang2025epona} & \gcmark & \rxmark & \gcmark & \gcmark & 7.50 & 82.80 \\

\midrule

BEVControl~\cite{yang2023bevcontrol} & \gcmark & \gcmark & \rxmark & \gcmark & 24.85 & - \\
BEVGen~\cite{swerdlow2024street} & \gcmark & \gcmark & \rxmark & \gcmark & 25.54 & - \\
MagicDrive~\cite{gao2023magicdrive} & \gcmark & \gcmark & \rxmark & \gcmark & 16.20 & - \\

\midrule

DriveDreamer-2~\cite{zhao2025drivedreamer} & \gcmark  & \gcmark&  \gcmark & \rxmark & 11.20 & 55.70 \\
OmniNWM~\cite{li2025omninwm} & \gcmark & \gcmark & \gcmark & \rxmark & 5.45 & 23.63 \\

\midrule

MagicDrive~\cite{gao2023magicdrive} & \gcmark & \gcmark & \gcmark & \gcmark & 18.75 & 218.12 \\
GenAD~\cite{zheng2024genad} & \gcmark & \gcmark & \gcmark & \gcmark & 15.40 & 184.00 \\
Panacea~\cite{wen2024panacea} & \gcmark & \gcmark & \gcmark & \gcmark & 16.96 & 139.00\\
Drive-WM~\cite{wang2024driving} & \gcmark& \gcmark & \gcmark & \gcmark & 15.80  &  122.70  \\
DriveDreamer-2~\cite{zhao2025drivedreamer} & \gcmark  & \gcmark&  \gcmark & \gcmark & 25.00 & 105.10 \\
MagicDrive-V2~\cite{gao2025magicdrive} & \gcmark & \gcmark&  \gcmark & \gcmark & 20.91 & 94.84 \\

\midrule

\modelname~\textit{(Single-shot)} & \gcmark & \gcmark & \gcmark & \rxmark & 8.60 & 46.99  \\
\modelname~\textit{(Single-shot)} & \gcmark & \gcmark & \gcmark & \gcmark & 13.24 & 84.71  \\
\modelname~\textit{(AR Gen.)} & \gcmark & \gcmark & \gcmark & \gcmark & 11.92 & 82.78 \\

\bottomrule
\end{tabular*}
\caption{
Quantitative comparison of multi-view video generation on nuScenes.
Methods are grouped according to whether they (i) avoid heavy volumetric intermediate representations (Vol.~Free),
(ii) support multi-view generation (MV),
(iii) generate video sequences (Vid.),
and (iv) operate without first-frame conditioning (FF~Free).
For AR generation we use 20 sampling steps per frame, while single-shot generation uses 30 sampling steps.
}
\label{tab_video}
\end{table}

\begin{table*}[t]
\centering
\small
\setlength{\tabcolsep}{3pt}
\begin{tabular*}{\textwidth}{@{\extracolsep{\fill}}cccccccccc@{}}
\toprule
\textbf{Steps} &
\textbf{KV Cache} &
\textbf{CFG-Free} &
\textbf{Cond. Cache} &
\textbf{FID} $\downarrow$ & \textbf{FVD} $\downarrow$ & \textbf{mAP} $\uparrow$ & \textbf{mIoU} $\uparrow$ &
\textbf{NPU Time} &
\textbf{GPU Time} \\
\midrule
1  & \gcmark  & \gcmark & \gcmark & 25.94 & 139.53 & 27.33 & 36.52 & 1.09~s/it & 0.84~s/it \\
2  & \gcmark  & \gcmark & \gcmark & 18.44 & 99.95 & 27.54 & 36.42 & 1.35~s/it & 0.98~s/it \\
3  & \gcmark  & \gcmark & \gcmark & 17.35 & 94.69 & 27.39 & 36.35 & 1.63~s/it & 1.15~s/it \\
4  & \gcmark  & \gcmark & \gcmark & 16.44 & 91.11 & 27.38 & 36.31 & 1.82~s/it & 1.28~s/it \\
5  & \gcmark  & \gcmark & \gcmark & 15.81 & 90.03 & 27.29 & 36.26 & 2.12~s/it & 1.43~s/it \\
\midrule
20 & \gcmark  & \gcmark & \gcmark & 11.92 & 82.78 & 26.96 & 36.25 & 5.75~s/it & 3.36~s/it \\
20 & \rxmark  & \gcmark  & \gcmark & 11.85 & 80.53 & 26.88 & 36.28 & 8.90~s/it & 8.15~s/it \\
20 & \rxmark  & \rxmark  & \gcmark & 12.01 & 86.10 & 28.13 & 36.33 & 17.15~s/it & 15.42~s/it \\
20 & \rxmark  & \rxmark  & \rxmark & 12.01 & 86.10 & 28.13 & 36.33 & 20.05~s/it & 15.94~s/it \\
\bottomrule
\end{tabular*}
\caption{
Inference configuration ablations. Latency~(s/it) is measured per autoregressive iteration generating six synchronized camera views on an NPU or H200 GPU. CFG denotes classifier-free guidance; Cond.~Cache denotes condition caching.
}
\label{tab:inference_ablation}
\end{table*}

\subsection{Ablation Studies}
\label{ablation}

\begin{figure}[t!]
    \centering
    \includegraphics[width=\columnwidth]{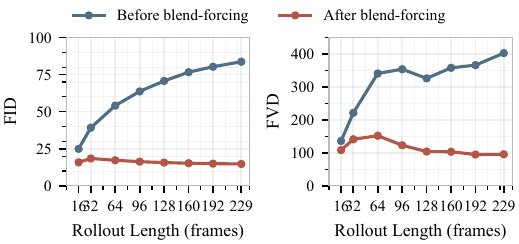}
    \caption{Ablation on the proposed blend-forcing on long-horizon generation.
We evaluate FID and FVD (lower is better) at different rollout lengths from 16 to 229 frames.}
    \label{fig:ablation_forcing}
\end{figure}

\begin{figure}[t!]
    \centering
    \includegraphics[width=\columnwidth]{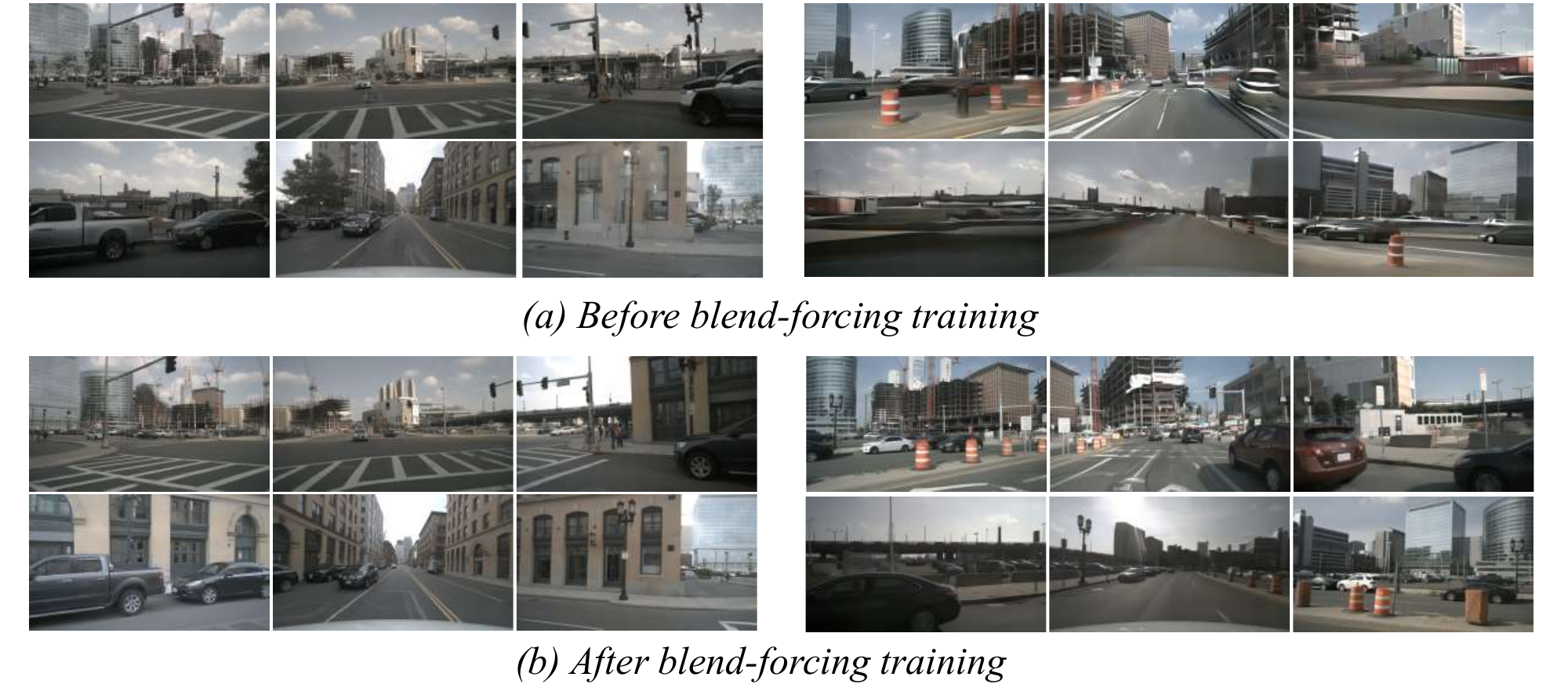}
    \caption{Qualitative comparison of long-horizon autoregressive generation before and after Blend-Forcing training. We generate a 229-frame six-view sequence and visualize the first frame (left) and the last frame (right). More cases can be found in Appendix~\ref{sec:app:qual}.}
    \label{fig:ablation_forcing_case}
\end{figure}

\textbf{Blend-Forcing Autoregressive Training.} 
Figure~\ref{fig:ablation_forcing} compares FID and FVD across autoregressive rollout lengths before and after blend-forcing. Without blend-forcing, both metrics steadily degrade as rollout length increases. In contrast, the post-training checkpoint exhibits more stable FID and FVD across the evaluated rollout lengths. The post-training checkpoint maintains image and video feature quality beyond the short training horizon. The qualitative comparison in Figure~\ref{fig:ablation_forcing_case} shows plausible initial frames from both models, with less structural degradation in the displayed final frames after blend-forcing.
Training rollouts contain up to 16 frames, with loss applied to the last eight, whereas evaluation extends to 229 frames. Self Forcing also supports video extrapolation through a rolling KV cache~\cite{selfforcing}; our results examine long-horizon rollout under six-view driving conditions. Sec.~\ref{sec:exp_setup} specifies how frame- and video-level metrics are computed at each rollout length.

\begin{figure}[t!]
    \centering
    \includegraphics[width=\columnwidth]{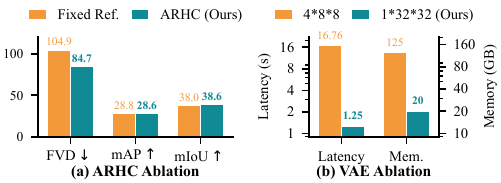}
    \caption{Ablation studies. (a)~ARHC vs.\ fixed reference-horizon conditioning on nuScenes. (b)~Autoencoder configuration comparison: per-iteration latency and GPU memory.}
    \label{fig:ablation_bars}
\end{figure}

\tit{Adaptive Reference-Horizon Conditioning}
To validate the effectiveness of ARHC, we compare it against a fixed reference-horizon strategy during the first training stage, where the model is always conditioned on three reference frames. As shown in Fig.~\ref{fig:ablation_bars}(a), ARHC achieves significantly lower FVD, while mAP and mIoU remain comparable.

\tit{Frame-Level Autoencoder}
We adopt a $1\times32\times32$ autoencoder with frame-level spatial compression instead of temporal VAEs (e.g., $4\times8\times8$) to support frame-level autoregressive generation.
Temporal VAEs inherently produce chunks of frames (e.g., 4 frames per latent step), making them unsuitable for frame-level AR iteration.
For an input of $T$ frames at resolution $H\times W$, the $4\times8\times8$ configuration yields $THW/256$ latent positions, whereas $1\times32\times32$ yields $THW/1024$. With otherwise matched tokenization, spatial-only compression therefore reduces the token count to one quarter. The corresponding quadratic full-attention term falls to one sixteenth, giving an idealized $16\times$ reduction for equal-duration inputs. End-to-end runtime also includes projections, feed-forward layers, autoencoder processing, and memory operations.
As shown in Fig.~\ref{fig:ablation_bars}(b), the $1\times32\times32$ configuration reduces per-iteration latency from 16.76s to 1.25s (a measured $13.4\times$ speedup) and GPU memory from 125\,GB to 20\,GB, making frame-level AR generation practically feasible.

\tit{Inference Efficiency Optimization} Table~\ref{tab:inference_ablation} analyzes the impact of sampling steps, classifier-free guidance (CFG~\cite{cfg}), and KV cache on both generation quality and inference latency. Our model generates six synchronized $576\times1024$ camera images per autoregressive step. On an H200 GPU, the three-step configuration takes 1.15 seconds per generated six-view frame, while one- and two-step configurations take 0.84 and 0.98 seconds with higher FID/FVD.
Reducing the number of denoising steps substantially accelerates inference while maintaining competitive visual and structural metrics, indicating that our model remains stable under significant step reduction.
Enabling KV cache reduces 20-step GPU latency from 8.15s to 3.36s, with FID changing from 11.85 to 11.92 and FVD from 80.53 to 82.78, while applying condition caching across diffusion steps further provides a consistent reduction in latency.
Using CFG (scale=2) improves mAP and slightly improves mIoU, but increases FID and FVD and nearly doubles GPU latency relative to scale=1 in the matched uncached-KV configurations. Therefore, avoiding CFG provides a better trade-off for low-latency rollout.
Overall, the results demonstrate that our framework supports flexible quality--efficiency trade-offs and can operate under low-latency constraints.

\section{Conclusion}
\label{sec:con}

We presented \modelname, a frame-level autoregressive multi-view generative framework for autonomous driving. Our approach combines a multi-view DiT with structured control, a two-stage training strategy (adaptive reference-horizon conditioning and blend-forcing), and system-level inference optimizations to achieve stable long-horizon generation with second-level inference latency. Experiments show strong reported FVD and structural metrics against the compared driving video generators, together with sustained rollout quality under recorded scene controls. We hope this work provides a practical step toward scalable, high-fidelity generative simulators for autonomous driving.

{\small
\bibliographystyle{ieeenat_fullname}
\bibliography{main}
}
\clearpage
\appendix
\twocolumn[{\centering\Large\bfseries Appendix\par\vspace{1em}}]
\label{supp:start}
\section{Model}
\label{sec:app:model}
Our model contains approximately \textbf{3.7B} parameters in total, including a pretrained generative backbone, a control branch for structured condition injection, and several lightweight embedding and projection modules.

\tit{Backbone MMDiT}
Our model is built upon a pretrained video generative backbone based on the MMDiT architecture~\cite{sd3}, containing approximately \textbf{2.5B} parameters.
The model follows a diffusion transformer design and supports multimodal conditioning, accepting text, image, and video inputs and generating video sequences in the latent space.
The backbone consists of 28 MMDiT blocks with a hidden dimension of 1792 and 14 attention heads.

\tit{Causal Conversion of Pretrained DiT}
To support autoregressive video generation, we convert the original bidirectional attention in the pretrained DiT into causal attention along the temporal dimension during the causal finetuning phase described in Sec.~\ref{app:train}. Specifically, tokens of each target frame are only allowed to attend to tokens from their current and previous frames, preventing information leakage from future frames during generation.

\tit{Cross-View Attention}
Let $\mathbf{h} \in \mathbb{R}^{B \times V \times T \times H' \times W' \times C'}$ denote the latent representation at a given DiT block, where $B$ is the batch size, $V$ the number of camera views, $T$ the temporal length, and $H', W', C'$ the spatial resolution and channel dimension, respectively. To enable temporal modeling, we reshape the tokens as $\mathbf{h}' \in \mathbb{R}^{B \times V \times (T H'W') \times C'}$ and apply standard self-attention over each video sequence independently for each camera view. Subsequently, to enable cross-view interaction, we reshape the tokens as $\mathbf{h}' \in \mathbb{R}^{B \times T \times (V H' W') \times C'}$ and apply self-attention over the panoramic spatial tokens from all views at each timestep. The cross-view attention mechanism in our model is inspired by DIVE~\cite{jiang2024dive}, which operates independently of temporal constraints and is unrelated to causal attention. Each cross-view attention block is initialized with the weights of the previous causal video block, and the output is zero-initialized to preserve the backbone logic at initialization. In our architecture, causal attention blocks and cross-view attention blocks are interleaved across layers as shown in Fig.~\ref{app:arch}, with one cross-view attention block inserted after every four backbone causal attention blocks, which allows the model to progressively learn both temporal dependencies and multi-view interactions in parallel. These modules introduce approximately \textbf{0.3B} additional parameters.

\tit{Control Branch Architecture}
We introduce an additional control DiT branch to inject structured control conditions into the generative process.
Empirically, we find that the control branch does not require the same depth as the backbone.
As demonstrated in Fig.~\ref{app:arch}, compared to 7 backbone MMDiT units, we use only 3 control MMDiT units in our model.
As a result, the control branch influences only the first few layers of the backbone through the control injection mechanism. The control branch contains approximately \textbf{0.8B} parameters.

In addition, several lightweight auxiliary modules contribute roughly \textbf{0.1B} parameters.
Overall, the full model contains approximately \textbf{3.7B} parameters.

\section{Data}
\subsection{Dataset Format}
We use nuScenes~\cite{caesar2020nuscenes} dataset for training and evaluation. After applying standard preprocessing steps commonly used in prior works~\cite{gao2023magicdrive}, each training sample contains the following components:

\begin{figure*}[!t]
    \centering
    \includegraphics[width=\textwidth]{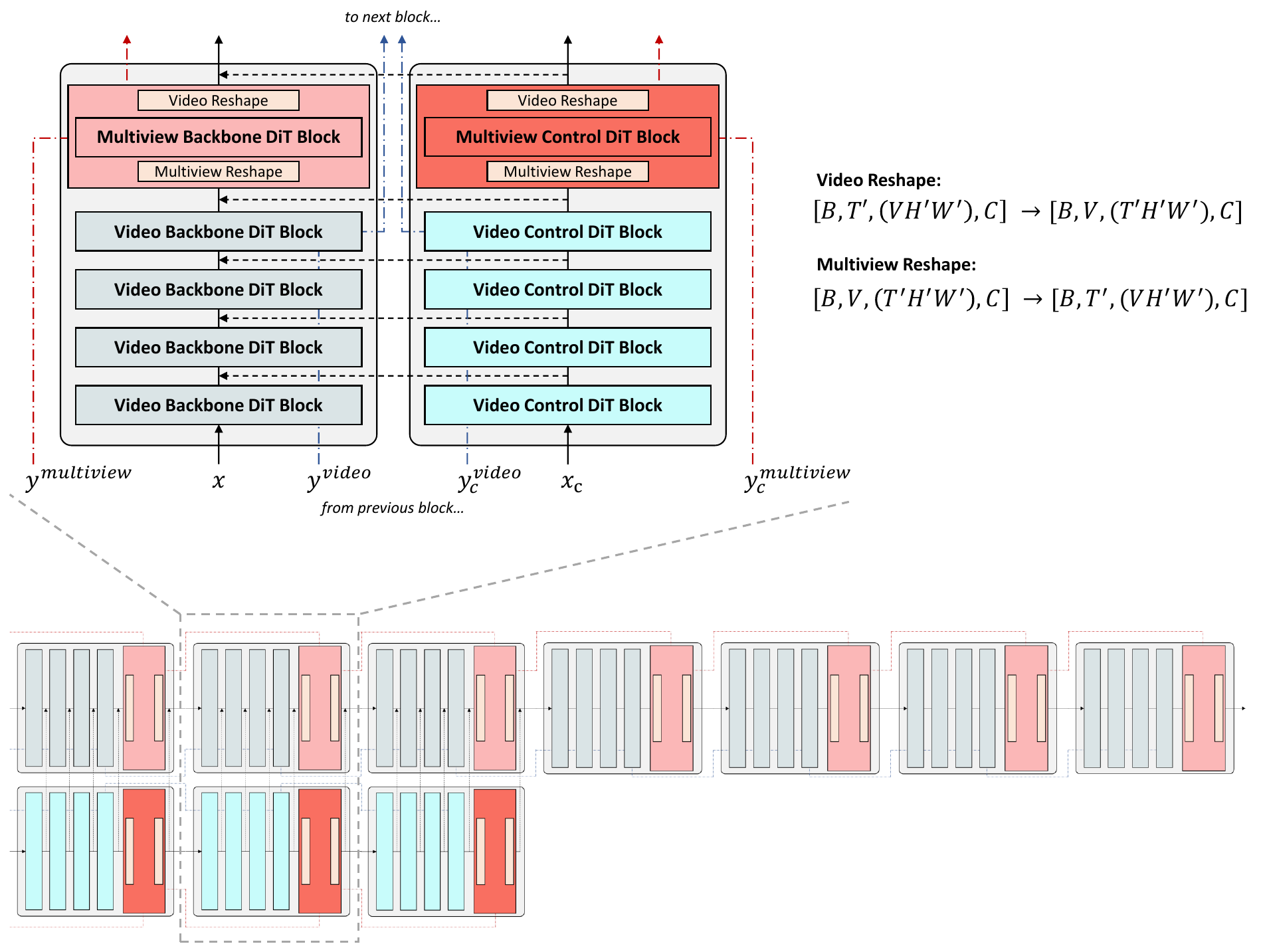}
    \caption{Our model architecture consists of 7 backbone units (each containing 4 backbone MMDiT blocks and 1 cross-view attention block) and 3 control units. We have provided a detailed description of a single backbone unit and its corresponding control unit. In this design, every four backbone blocks are alternately arranged with a cross-view attention block.}
    \label{app:arch}
\end{figure*}

\tit{Multi-view Video}
We sample 16-frame training clips from the six synchronized camera streams of each scene.
The RGB images are encoded into a latent representation using a pretrained autoencoder, which serves as the input and prediction target of the DiT.
The latent tensor therefore represents a temporally uncompressed and spatially compressed version of the original multi-view video.

\tit{Camera Parameters}
Each camera is associated with its intrinsic and extrinsic parameters.
The camera intrinsics describe the projection from camera coordinates to the image plane, while the extrinsics specify the camera pose with respect to the ego vehicle and global coordinate system through a rotation and translation.
These parameters are used to maintain geometric consistency across camera views.

\tit{BEV Semantic Map}
Bird's-eye-view (BEV) semantic map of the environment centered at the ego vehicle.
The BEV map contains multiple binary channels representing different map elements such as road regions and pedestrian crossings.
This representation provides a compact spatial description of the scene layout.

\tit{Bounding Boxes}
Dynamic agents in the scene are represented using 3D bounding boxes together with object category labels.
Each box is represented by its 3D pose and size and can be converted to its eight vertices.
Additional masks indicate whether a box corresponds to a valid object or padding.
The categories cover common traffic participants such as vehicles, pedestrians, bicycles, and traffic cones.

\tit{Text Caption}
Each scene is associated with a natural language description summarizing the driving scenario.
During training, the captions are encoded using a pretrained text encoder to obtain token-level embeddings, which are used as conditioning inputs for the generative model.

\tit{Ego-Motion}
We represent the ego vehicle motion using a sequence of relative ego poses between frames.
Each pose consists of a 3D rotation and translation describing the transformation from the initial coordinate frame to the current frame.
This motion representation is used to capture the temporal evolution of the camera rig.

\subsection{Data Processing}
Following common practices in diffusion-based video generation, we encode images and text into latent representations before training.
For text conditioning, we use the T5-XXL text encoder from the T5 family~\cite{raffel2020exploring}.
The input captions are tokenized and converted into token-level embeddings which are used as conditioning inputs for the DiT backbone.
For visual inputs, we adopt the $1\times32\times32$ autoencoder from the DC-AE~\cite{chen2024deep} implementation provided in Diffusers.
All images are first resized to a resolution of $576\times1024$, preserving the original aspect ratio of the nuScenes images ($900\times1600$), so no cropping is required.
The resized frames are then encoded into latent representations using the autoencoder.
After encoding, each frame is represented as a latent feature map with spatial resolution $18\times32$, which serves as the input and prediction target of the diffusion transformer during training and inference.
In addition to autoencoder encoding, we project the BEV road layout and 3D object bounding boxes into each camera view using annotations provided by the nuScenes dataset together with the camera calibration parameters like lidar projections, following a standard projection pipeline~\cite{yang2025drivearena}.

\begin{figure*}[!t]
    \centering
    \includegraphics[width=\textwidth]{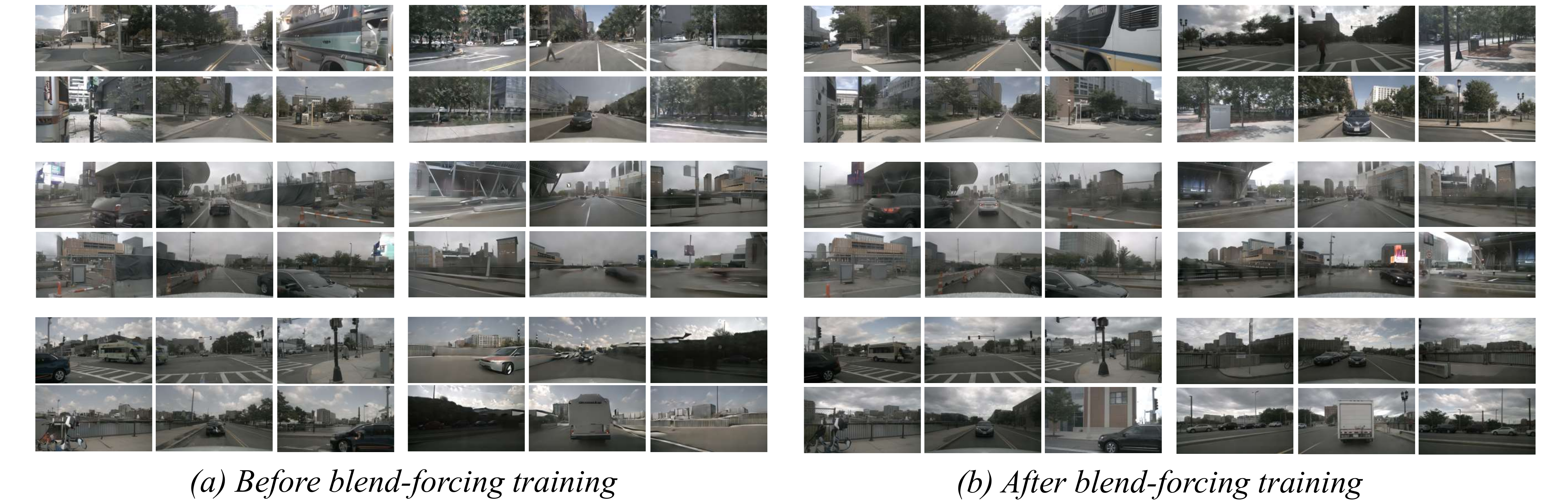}
    \caption{Qualitative comparison of long-horizon autoregressive generation before (part a) and after (part b) blend-forcing training. We generate a 229-frame six-view sequence and visualize the first frame (left of each part) and the last frame (right of each part).}
    \label{app:case}
\end{figure*}

\section{Training}
\label{app:train}

Table~\ref{tab:training_schedule} summarizes the optimization phases, and Sec.~\ref{sec:exp_setup} gives the training configuration. We provide further details on reference sampling and autoregressive training below.

\tit{Adaptive Reference-Horizon Conditioning}
For a training sequence of 16 frames, we condition the generation of future frames on a variable number of preceding reference frames.
Specifically, a reference length from 0 to 15 can be sampled to condition the prediction of subsequent frames.
The input latents therefore consist of a mixture of clean reference-frame latents and noisy target-frame latents.
We design a sampling strategy to select the number of reference frames with a biased distribution, giving higher probability to short horizons (e.g., 1-3 frames), which are more common in autoregressive rollout.

\tit{Blend-Forcing Autoregressive Training}
During training, the blending coefficient $\alpha$ increases linearly over time to gradually transition from ground-truth conditioning to model-generated conditioning. The exact growth rate of $\alpha$ is set to $1\times10^{-4}$ thus will be capped at 1 after 10,000 steps, while the reported training run ends around 7,000 BF steps, at which point $\alpha\approx0.7$.
To keep the model generation quality stable, we perform a step of standard flow matching~\cite{flowmatching} training as described in Sec.~\ref{subsec:training} for every step of blend-forcing.
To improve the robustness, we randomly increase the length of the generated sequence up to 16 frames to increase the level of degradation on the generated conditioning data during training, while still using only the last 8 frames for the BF loss.

\section{Theoretical Analysis of Blend-Forcing}
\label{app:blend_theory}

We analyze how joint supervision on clean and blended histories relates to prediction under generated history. The analysis follows a fixed time-aligned history pair and the conditional flow-matching objective used in training.

\subsection{Setup and Local Smoothness}
Let $h$ be the concatenation of clean reference latents and $h+e$ the corresponding generated history. Fix the target latent $\mathbf{Z}_{i,\tau}$, flow time $\tau$, controls $\mathbf{C}_i$, and velocity target $y=\mathbf{U}^{*}_{i,\tau}$. Write
\begin{equation}
F(h')=\mathbf{u}_\theta(\mathbf{Z}_{i,\tau},\tau\mid h',\mathbf{C}_i),
\qquad f(s)=F(h+se).
\end{equation}
We use Euclidean norms on concatenated latents and outputs, and the induced operator norm on Jacobians. Assume $F$ is continuously differentiable along the history segment, with Jacobian $J$ satisfying
\begin{equation}
\|J(h+se)-J(h+te)\|_{\mathrm{op}}
\le M|s-t|\|e\|,
\quad s,t\in[0,1].
\end{equation}
This assumption bounds the local variation of the predictor's response to history perturbations. Since $f'(s)=J(h+se)e$, it implies
\begin{equation}
\|f'(s)-f'(t)\|\le M|s-t|\|e\|^2.
\label{eq:directional_smoothness}
\end{equation}
The prediction residual $r_s=f(s)-y$ uses the same target at every position on the segment. Thus $r_0$ and $r_\alpha$ correspond to the clean and blended conditioning cases supervised by standard FM and BF.

\subsection{Generated-History Residual Bound}
For $0<\alpha\le1$, define the secant extrapolation remainder
\begin{equation}
\Delta_\alpha=f(1)-f(0)-\frac{f(\alpha)-f(0)}{\alpha}.
\end{equation}
The fundamental theorem of calculus and a change of variables give
\begin{equation}
\Delta_\alpha=\int_0^1\left[f'(t)-f'(\alpha t)\right]\,dt.
\end{equation}
Using Eq.~\ref{eq:directional_smoothness}, we obtain
\begin{equation}
\begin{aligned}
\|\Delta_\alpha\|
&\le\int_0^1 M(1-\alpha)t\|e\|^2\,dt\\
&=\frac{M}{2}(1-\alpha)\|e\|^2.
\end{aligned}
\end{equation}
Subtracting the common velocity target gives the exact residual decomposition
\begin{equation}
r_1=\frac{r_\alpha}{\alpha}
-\frac{1-\alpha}{\alpha}r_0+\Delta_\alpha.
\end{equation}
The triangle inequality yields Eq.~\ref{eq:bf_transfer}. For an affine response along the segment, $\Delta_\alpha=0$, and the clean and blended residuals determine the generated-history residual exactly.

\subsection{Connection to Paired Training}
The standard FM and BF objectives minimize squared prediction residuals under clean and blended histories, respectively. The bound links these supervised quantities to the residual at the generated-history endpoint. It also explains the roles of both objectives: clean-history supervision maintains the reference prediction, while blended-history supervision trains the response in the direction of model-generated deviations.

The schedule controls both perturbation magnitude and proximity to the generated-history endpoint. Early in training, $\|h+\alpha e-h\|=\alpha\|e\|$ keeps the conditioning perturbation small. As $\alpha$ increases, the factors $1/\alpha$, $(1-\alpha)/\alpha$, and $(1-\alpha)$ all decrease. At $\alpha=0.7$, the bound becomes
\begin{equation}
\|r_1\|\le\frac{10}{7}\|r_{0.7}\|
+\frac{3}{7}\|r_0\|+0.15M\|e\|^2.
\end{equation}
This provides a local account of the training curriculum: progressively stronger generated-history perturbations are supervised together with the clean reference, bringing the training conditions closer to those used in autoregressive generation.

\section{Evaluation}
\label{app:evaluation}
The metric definitions and long-horizon evaluation procedure are provided in Sec.~\ref{sec:exp_setup}. We use the inference configurations specified in the main paper for each comparison.

\section{Additional Qualitative Results}
\label{sec:app:qual}

In addition to the examples presented in Sec.~4.3, we provide additional qualitative results in Fig.~\ref{app:case} covering more scenarios.

\section{Scope and Limitations}

Our work focuses on training a multi-view video generator for sustained frame-level autoregressive generation under structured scene conditions. The experiments evaluate generation quality, long-horizon rollout, and inference efficiency using recorded nuScenes conditions. Closed-loop interaction with a driving policy has not been evaluated in this work. Connecting policy predictions to a scene-state update mechanism provides a natural next step toward policy-driven visual simulation.

Within the evaluated generation setting, a typical failure case involves blurred rendering of roadside trees. This may arise from the loss of high-frequency details under the high-compression autoencoder. The model may also struggle with extreme weather conditions and long-tailed scenarios, motivating further improvements in visual detail and scenario coverage.

\section{Discussion and Future Directions}

Future work can extend rollout duration through stronger temporal modeling and improved autoregressive training. Distillation and model compression offer routes toward real-time generation (\eg, around 10 frames per second). Integrating the generator with a driving agent and a scene-state update mechanism would enable closed-loop policy training and evaluation. A further question is how policies trained in such generated environments transfer to real driving.

\section{Attention Cost and Measured Latency}
\label{app:efficiency}
The latent token counts and quadratic attention cost comparison are given in the frame-level autoencoder ablation in Sec.~\ref{ablation}. The measured latency ratio in Fig.~\ref{fig:ablation_bars}(b) is $16.76/1.25\approx13.4$.

\end{document}